%% file: egpaper_for_review.tex
\documentclass[10pt,twocolumn,letterpaper]{article}

\usepackage{iccv}
\usepackage{times}
\usepackage{epsfig}
\usepackage{graphicx}
\usepackage{amsmath}
\usepackage{amssymb}

\usepackage{floatrow}
\usepackage{verbatim}
\usepackage{amssymb,amsmath,amsthm}
\usepackage{subfigure}

\usepackage{xspace}

\usepackage{enumitem}
\usepackage{multirow}
\usepackage{bm}
\usepackage{sidecap}
\usepackage{xspace}
\newcommand*{\our}{ADAGE\@\xspace}

\usepackage[pagebackref=true,breaklinks=true,letterpaper=true,colorlinks,bookmarks=false]{hyperref}

\iccvfinalcopy 

\def\iccvPaperID{8} 

\ificcvfinal\fi

\begin{document}

\title{Hallucinating Agnostic Images to Generalize Across Domains}

\author{Fabio M. Carlucci$^{1}$\thanks{This work was done while at University of Rome Sapienza, Italy}
\hspace{0.8cm}
Paolo Russo$^{2}$
\hspace{0.8cm}
Tatiana Tommasi$^{3}$  
\hspace{0.8cm}
Barbara Caputo$^{3,4}$ \vspace{2mm}\\ 
\hspace{0.8cm}
$^{1}${Huawei Noah's Ark Lab}, London
\hspace{0.5cm}
$^{2}$University of Rome Sapienza, Italy\vspace{1mm}\\
$^{3}$Politecnico di Torino, Italy \hspace{0.5cm}
$^{4}$Italian Institute of Technology\vspace{2mm}\\
{\tt\small fabio.maria.carlucci@huawei.com \hspace{0.2cm} prusso@diag.uniroma1.it}\\
{\tt\small \{tatiana.tommasi, barbara.caputo\}@polito.it}
}

\maketitle
\ificcvfinal\thispagestyle{empty}\fi

\begin{abstract}
\input{abstract_dream.tex}
\vspace{-0.4cm}
\end{abstract}

\section{Introduction}
\label{sec:introduction}
\input{introduction}

\section{Related Work}
\label{sec:related}
\input{related_work.tex}

\section{Agnostic Domain Generalization}
\input{architecture}

\label{sec:architecture}
\section{Experiments}
\label{sec:experiments}
\input{expers_tat.tex}

\section{Conclusions}
This paper proposes the first end-to-end joint image- and feature-level adaptive solution for DG.
We define a new network, named \our, able to hallucinate domain agnostic images guided by two 
adversarial adaptive conditions at pixel and feature level. \our can be seamlessly used both for DG and multi-source unsupervised DA: it achieves impressive results on several benchmarks, outperforming the current state of the art by a significant margin. We plan to extend \our also to the open set multi-source DA and DG scenarios.

\section*{Appendix}
\label{sec:appenix}
\input{supplementary.tex}

{\small
\bibliographystyle{ieee}
\bibliography{egbib}
}

\end{document}

%% file: abstract_dream.tex
The ability to generalize across visual domains is crucial for the robustness of artificial recognition systems. Although many training sources may be available in real contexts, the access to even unlabeled target samples cannot be taken for granted, which makes standard unsupervised domain adaptation methods inapplicable in the wild. In this work we investigate  how to exploit multiple sources by hallucinating a deep visual domain composed of images, possibly unrealistic, able to maintain categorical knowledge while discarding specific source styles.
The produced agnostic images are the result of a 
deep architecture that applies pixel adaptation on the original source data guided by two adversarial domain classifier branches at image and feature level. 
Our approach is conceived to learn only from source data,  but it seamlessly extends to the use of unlabeled target samples. Remarkable results for both multi-source domain adaptation and domain generalization support the power of hallucinating agnostic images in this framework.

%% file: introduction.tex
Domain Adaptation (DA) is at its core the quest for principled algorithms enabling the generalization of visual recognition methods. Given at least a source domain for training, the goal is to obtain recognition results as good as those achievable on source test data on \emph{any}  other target domain, in principle belonging to a different probability distribution.
While originally defined assuming
to have access to annotated data from a single source domain, and to unlabeled data from a different target domain \cite{Saenko:2010}, there is growing interest on how to leverage over multiple sources,
and for domain generalization (DG), i.e. the case when it is not possible to access target data of any sort a priori.
Algorithm-wise, 
three strategies have been proposed, i.e. 
dealing with model \cite{DinnocenteGCPR2018, MLDG_AAA18}, feature \cite{LongZ0J17,PAMI2018salzmann}, or image adaptation \cite{russo17sbadagan,cycada}.  
A basic assumption for both feature and image adaptation approaches is the existence of 
a shared space among domains, however only feature-based methods attempt to explicitly 
identify it \cite{hoffman_eccv12,JhuoLLC12,Bousmalis:DSN:NIPS16}. In the image-based approaches, the domain generic component is always silently recombined with 
the specific domain style to obtain images that show the same content of the target, but with source-like 
appearance or vice-versa \cite{cycada,russo17sbadagan,liu2017unsupervised}. 
Moreover, although these 
methods have shown to be effective in the single source scenario , it is 
questionable whether they could be extended to multi-source DA, 
or to DG. 

With this paper we make two contributions: (1) we introduce image adaptation for DG,
(2) we propose an architecture that exploits the power of layer aggregation to hallucinate samples of the latent pixel space shared among domains. 
We call our method Agnostic DomAin GEneralization (ADAGE). 
To our knowledge it is the first solution 
to introduce an image-level component in an end-to-end deep learning architecture for DG and that can work seamlessly also in the multi-source unsupervised DA setting. 

We start by acknowledging that the notion of visual cross-domain generic information is intuitive yet ambiguous, as ground 
truth examples of pure semantic images without a characteristic style do not exist. Thus, while it is possible to interpret the produced samples as capturing domain agnostic knowledge, it should be clear that they are built for the network's benefit only and we do not expect them to be pleasant to the human eye.
Practically, we let the network learn what this generic information is through a mapping
guided by adversarial adaptive constraints. These constraints are applied directly on the agnostic space, 
rather than on standard images that always contain domain-specific information. 

To realize the mapping we define a dedicated convolutional structure 
loosely related to a previous image colorization
network \cite{carlucci2018text}. The new architecture has a low  
number of parameters which prevents overfitting and at the same time allows to 
comfortably accommodate two gradient reversal layers that adversarially exploit both image and feature
classification across domains.  As the image domain discriminator maintains the ability to evaluate the 
similarity of a target image to the different source domains, it is straightforward to extend the method to multi-source DA, and learn how to bias the classification loss towards the sources that are more similar to the target. 

We test \our in the DG and multi-source DA scenarios, comparing against 
recent approaches \cite{MLDG_AAA18,MDAN_ICLRW18,cocktail_CVPR18}. In all experiments, for both settings, \our significantly outperforms the state of the art. An ablation study and visualizations of the agnostic domain images complete our experimental study.

%% file: related_work.tex
In \textbf{single source DA}, \emph{feature adaptation} approaches  
aim at learning deep domain invariant representations 
\cite{LongZ0J17,Sun:CORAL:AAAI16,carlucci2017auto,carlucci2017just,PAMI2018salzmann,WenLi:ECCV2016,TRUDA-NIPS16_savarese, haeusser17,saito2017asymmetric}.
Other methods
rely on adversarial loss functions \cite{Ganin:DANN:JMLR16,Tzeng_ICCV2015,sankaranarayanan2017generate}. 
Besides end-to-end trained architectures also two-step adaptive networks have shown practical advantages \cite{Hoffman:Adda:CVPR17,LOAD_ICRA}.
Most of work based on \emph{image adaptation} aims at producing either 
target-like source images or source-like target images, but it has been recently shown that integrating both the
transformation directions is highly beneficial \cite{russo17sbadagan,cycada}. In particular 
\cite{cycada} combines both image and feature-level adaptation. 
Considering that the proposed network contains two generators, three discriminators and one classifier for 
a single source-target domain pair, its extension to multi-source DA,  and even more to DG, is not straightforward.
\textbf{Multi-source DA} was initially studied from a theoretical point of view
\cite{Crammer_JMLR08}.
Within the context of convnet-based approaches,
the vanilla solution of collecting all the source data in a 
single domain is already quite effective. 
Only very recently two methods presented multi-source deep learning approaches that improve over this 
baseline. The method proposed in \cite{cocktail_CVPR18} builds over \cite{Ganin:DANN:JMLR16}
by replicating the adversarial domain discriminator branch for each available source.
A similar multi-way adversarial strategy is used also in \cite{MDAN_ICLRW18}, but this work comes 
with a theoretical support that frees it from the need of 
learning the source weights.

\begin{figure*}[!t]
\centering \vspace{-2mm}
\includegraphics[width=0.9\textwidth]{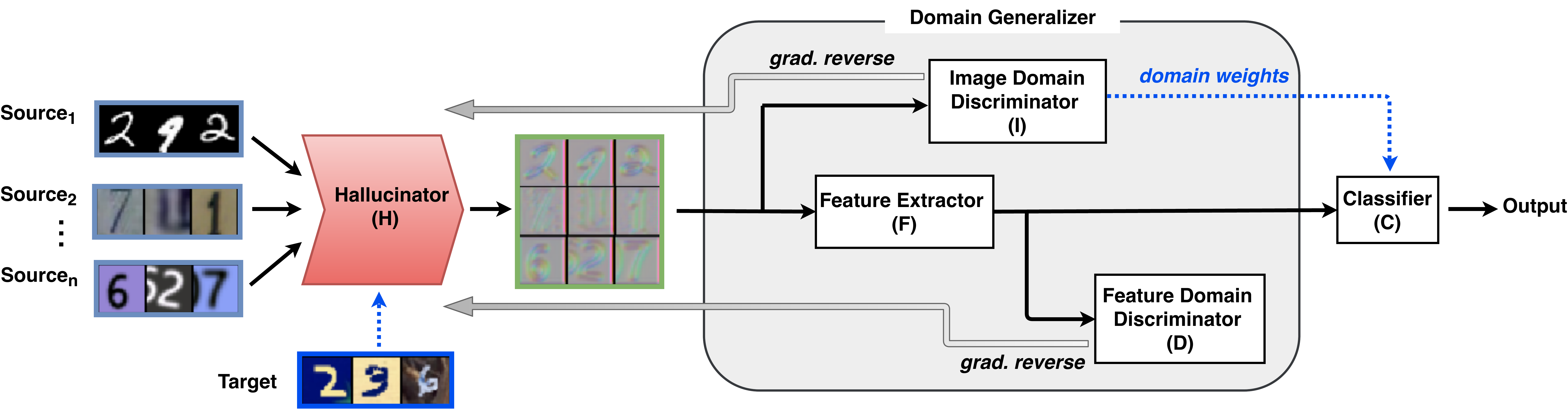}
\caption{A schematic description of \our. 
All samples (including target ones, in the DA setting) follow the same path in the network. 
The inverted gradient from $I$ flows through $H$ driving image modifications towards
domain confusion. Similarly, the gradient from $D$ also inverted, is backpropagated through $F$ and $H$
so that both the feature and the image dedicated blocks benefit from a further push towards the domain agnostic
space. The classification gradient travels through the whole network, excluding $I$ and $D$.}
\label{fig:architecture}
\vspace{-1mm}
\end{figure*}

In the \textbf{DG setting}, no access to the target data is allowed, thus the
main objective is to look across multiple sources for shared 
factors  
which
are either searched at \emph{model-level} to regularize the learning process on the sources, 
or at \emph{feature-level} to learn some domain-shared representation. 
Deep model-level strategies are presented in \cite{MassiRAL,hospedalesPACS,DinnocenteGCPR2018}. 
The first work proposes a weighting procedure on the source models, while the others aim at separating 
the source knowledge into domain-specific and domain-agnostic sub-models either with a low-rank parametrized network
or through a dedicated learning architecture with a shared backbone and source-specific aggregative modules.
A meta-learning approach was recently presented in \cite{MLDG_AAA18}.
Regarding feature-based methods, \cite{doretto2017} proposed to exploit a Siamese architecture
to learn an embedding space where samples from different source domains but same labels are 
projected nearby, while samples from different domains and different labels are mapped far
apart. Both works \cite{DGautoencoders,Li_2018_CVPR} exploit deep  autoencoders
for DG still focusing on representation learning.
New DG approaches based on \emph{data augmentation} have shown promising results.
Both  \cite{DG_ICLR18} and \cite{MurinoNIPS2018} propose domain-guided perturbation of the 
input instances in the embedding space, with the second work able to generalize to new targets
also when starting from a single source. 

Although a two-step DG solution involving an image-adaptive process, followed by a deep classifier with feature adversarial training is always possible \cite{wacv19}, we go beyond this na\"ive strategy. 
Differently from GAN-based methods that need a typical alternating
training between image adaptation and classification, we train the whole model of \our with a single optimizer while performing adversarial training by inverting the gradient originating from two domain discriminators at image and feature level.

%% file: architecture.tex
We assume to observe $i=1\ldots S$ source domains with the $i$th domain containing $N_i$ 
labeled instances $\{x_j^i,y_j^i\}_{j=1}^{N_i}$, where $x_j^i$ is the $j$th 
input image and $y_j^i\in\{1\ldots M\}$ is the class label. 
In addition we also have an unlabeled target domain whose data $\{x_j^{t}\}_{j=1}^{N_t}$ might 
(DA) or might not (DG) be provided at training time. All the source and target domains
share the same label space, 
but their marginal distribution is different thus inducing a domain shift. 
The goal of \our is to achieve domain generalization by hallucinating images stripped down of domain specific information, that thus can be seen as samples of a machine-created agnostic domain. 
We obtain this by learning to modify the images such that it becomes impossible to identify their original
source domain both from their pixels and from the extracted features, while maintaining their relevant semantic 
information.
Figure \ref{fig:architecture} shows our architecture, consisting of two main components: (1) the \emph{Hallucinator} block, in charge of generating the agnostic images from the input samples, and (2) the \emph{Domain Generalizer}, that performs adaptation from the new domain. 
The architecture is end-to-end, meaning that the two components are interconnected and trained jointly.

\begin{figure}[!hb]
\centering
\includegraphics[width=0.95\textwidth]{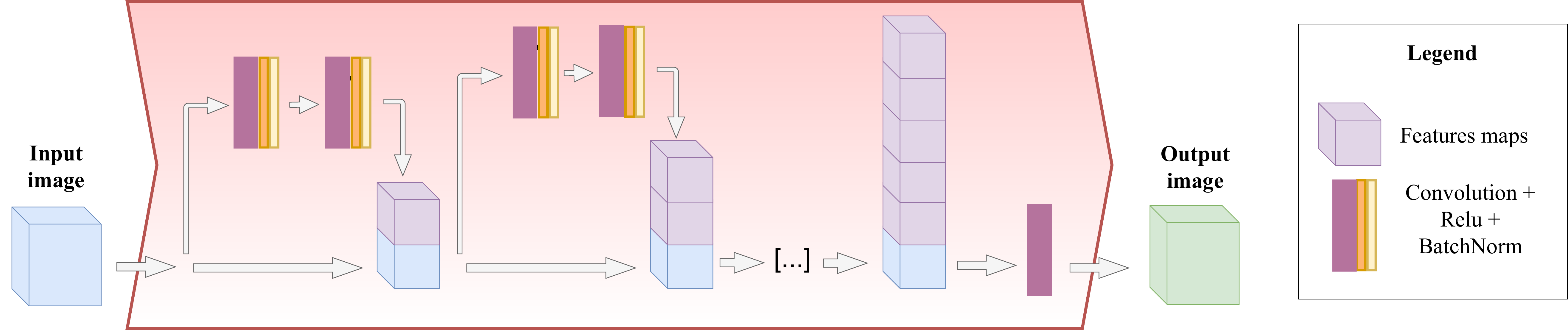}
\caption{The Hallucinator. 
The output of the two multicolor blocks (Convolutional + Relu + Batch Normalization) are concatenated with the previous inputs, forming a group of images and features maps that grow along the depth of the network. The number of features increases from 3 (input data) to 256 (final aggregation step), while a last Convolutional layer squeezes the features back into 3 channels, interpretable as an RGB image. } 
\label{fig:transformer}
\centering
\end{figure}

The \textbf{Hallucinator (H)} modifies the input images to remove their domain-specific style. To achieve this, we got inspiration from the colorization literature and define a new structure exploiting the power of layer aggregation \cite{shelhamerdeep}: the output of two $3\times 3$ convolutional layers, each followed by Relu and Batch Normalization 
are stacked up with the input and propagated to every subsequent layer (see Figure \ref{fig:transformer}). Specifically, the produced feature build up in size resulting in a growing sequence of 
$\{3,8,16,32,64,128\}$ maps, after which a convolution layer brings them down to 3 channels, 
interpretable as RGB images. 
With respect to previous mapping architectures proposed within the context of depth colorization \cite{carlucci2018text},
our hallucinator has a significantly lower number of parameters thanks to its incrementally aggregative structure.
This is crucial for generalization both because it reduces the risk of overfitting to the available sources
and because leaves space for building a multi-branch following network able to impose constraints that  
in turn will lead to 
learning a stronger and more stable hallucinator in our end-to-end framework.

The \textbf{Domain Generalizer} is composed by the \emph{Image Domain Discriminator} $I$, 
the \emph{Feature Domain Discriminator} $D$ and the \emph{Feature Extractor} $F$. The first two impose 
respectively an adversarial generalization condition on the pixels and on the feature extracted from the images
produced by $H$, while the third defines an intermediate step between the first two. Moreover, thanks to its
direct connection with the \emph{Classifier} $C$, it allows to maintain the basic semantic knowledge in the
hallucinated images, so that, despite they lack domain style, their label can still be recognized.

The Image Domain Discriminator $I$ receives as input the images produced by $H$
and predicts their domain label. More in details, this module is a multi-class classifier 
that learns to distinguish among the $S$ source domains in DG, and $S+1$ in DA 
(including the target), by minimizing a simple cross-entropy loss $\mathcal{L}_{I}$.
The information provided by this module is used in two ways:  to adversarially 
guide the hallucinator $H$ to produce images with confused domain identity, and
to estimate a similarity measure between the source and the target data when available.
The first task is executed through a gradient reversal layer as in \cite{Ganin:DANN:JMLR16}.
The second is obtained as a byproduct of the domain classifier $I$ by collecting the
probability of every source sample in each batch to be recognized as belonging to the target.

The Feature Domain Discriminator $D$ is analogous to $I$ but, instead of images,
it takes as input their features, performing domain classification by minimizing the cross-entropy
loss $\mathcal{L}_{D}$. During backpropagation, the inverted gradient regulates the feature 
extraction process to confuse the domains.
Finally, the Feature Extractor $F$, as well as the  Classifier $C$, is a standard deep learning module. 
We built both of them with the same network structure used in \cite{MDAN_ICLRW18} to 
put them on equal footing. In particular, in the DG setting the classifier learns to 
distinguish among the $M$ categories of the sources by minimizing the cross-entropy 
loss $\mathcal{L}_{C}$, while for the DA setting it can also provide the classification 
probability on the target samples $p(x^t)=C(F(H(x_t)))$ that is used to minimize the 
related entropy loss $\mathcal{L}_{E}=p(x^t)log(p(x^t))$. 

If we indicate with $\theta$ the network parameters and we use subscripts to identify the
different network modules, we can write the overall loss function optimized 
by \our as:
\vspace{-1mm}
\begin{align}
 \mathcal{L}(\theta_H,\theta_F,&\theta_D,\theta_{I},\theta_C) =   \nonumber \\ 
\sum_{i=1}^{S,S+1}\sum_{j=1}^{N^i} ~~& \mathcal{L}_C^{j,i}(\theta_H,\theta_F,\theta_C) +  
\eta \mathcal{L}_{E}^{j,i=S+1}(\theta_H,\theta_F,\theta_C)  \nonumber\\ 
      & - \lambda \mathcal{L}^{j,i}_D(\theta_H,\theta_F,\theta_D) - \gamma \mathcal{L}^{j,i}_I(\theta_H,\theta_I)~. 
 \vspace{-2mm}
\label{eq:main_loss}
\end{align}

We remark that, as specified by its superscripts, $ \mathcal{L}_{E}^{j,i=S+1}$ is only active in the DA setting, 
while $\mathcal{L}_D$ and $\mathcal{L}_I$ in the DA case deal with an $\{S+1\}$-multiclass task involving also
the target together with the source domains.

As can be noted from (\ref{eq:main_loss}), the number of meta-parameters of our approach is very limited.
For $\lambda$ we use the same rule introduced by \cite{Ganin:DANN:JMLR16} that grows the importance of the
feature domain discriminator with the training epochs: 
$\lambda_k = \frac{2}{1+exp(-10k)}-1$, where $k = \frac{current\_epoch}{total\_epochs}$. 
We set $\gamma_k=0.1 \lambda_k$ so that only a small portion of the full gradient of the image 
domain discriminator is backpropagated: in this way we can still get useful similarity measures 
among the domains while progressively guiding the hallucinator to make them alike. 
When the image adaptation part is enough to close the domain gap, the feature discriminator 
loss might be abnormally high causing divergence. We easily obviate such extreme cases by maintaining a 
record on the initial feature discriminative loss and avoiding the loss backpropagation if it is higher 
than twice its initial value. 
Finally, the experimental evaluation indicates that \our is robust to the exact choice of $\eta$, thus we keep 
it always fixed to 0.5 just for simplicity.

%% file: expers_tat.tex
\begin{table*}[ht]
    \centering
\resizebox{0.9\textwidth}{!}{
\begin{tabular}{@{}l@{~~}l@{}c@{}c@{}c@{}c}  
\hline
   \multicolumn{2}{c}{ \multirow{3}{*}{Sources}}& SVHN & SVHN & MNIST-M&\multirow{4}{*}{Avg.}\\ [0.82ex]
  \multicolumn{2}{c}{ }& MNIST-M & MNIST & SYNTH\\ [0.82ex]
   \multicolumn{2}{c}{ }& SYNTH & SYNTH & MNIST\\ [0.82ex] \cline{2-5}
  \multicolumn{2}{c}{Target} & MNIST & MNIST-M & SVHN\\ \hline
 \multirow{3}{*}{DG}  & combine sources             &  98.7  & 62.6 & 69.5 & 76.9\\ 
  & MLDG \cite{MLDG_AAA18}   &  99.1  & 61.2 & 69.7 & 76.7\\\cline{2-6}
 & \our          &  99.1  & \textbf{66.3} & \textbf{76.4}  & \textbf{80.3}\\
 \hline \hline
  \multirow{4}{*}{DA} & combine sources &  98.7  & 62.6 & 69.5 & 76.9\\ 
   & combine DANN \cite{MDAN_ICLRW18}   &  92.5   & 65.1  &  77.6 & 78.4\\
   & MDAN \cite{MDAN_ICLRW18}   &  97.9   &  68.7 &  81.6 & 82.7\\ \cline{2-6}
   & \our     &  \textbf{99.3}   &  \textbf{88.5} &  \textbf{86.0} & \textbf{91.3}\\
\hline
\end{tabular}
\hspace{1.5cm}
\begin{tabular}{@{}l@{~~}l@{}c@{}c@{}c}
\hline
   \multicolumn{2}{c}{ \multirow{4}{*}{Sources}}& SYNTH & SYNTH &\multirow{5}{*}{Avg.}\\
   \multicolumn{2}{c}{ }& MNIST & MNIST &\\
   \multicolumn{2}{c}{ }& MNIST-M & SVHN &\\
   \multicolumn{2}{c}{ }& USPS & USPS &\\ \cline{2-4}
  \multicolumn{2}{c}{Target} & SVHN & MNIST-M &\\ \hline
   \multirow{3}{*}{DG}  & {combine sources}             &  73.2  &  61.9& 67.5\\ 
  & MLDG \cite{MLDG_AAA18}     &  68.0  &  65.6 &  66.8 \\ \cline{2-5}
& \our  &  \textbf{75.8}  &  \textbf{67.0} & \textbf{71.4} \\
 \hline \hline
 \multirow{4}{*}{DA} & {combine sources}             &  73.2  &  61.9 & 67.5\\ 
  & combine DANN \cite{cocktail_CVPR18}   &  68.9   & 71.6  & 70.3 \\
  & DCTN \cite{cocktail_CVPR18}   &  77.5   &   70.9 & 74.2\\\cline{2-5}
  & \our     & \textbf{85.3}    &  \textbf{85.3}  & \textbf{85.3} \\
 \hline
\end{tabular}
}
 \caption{Classification accuracy results on the digits images \emph{Left}: experiments with three sources. \emph{Right}: experiments with four sources.}\label{table:34sources}
\end{table*}

\begin{figure*}[t]
    \centering\vspace{-3mm}
\subfigure[MNIST]{\label{fig:mnist_adaptation_target_mnistm}\includegraphics[width=0.21\textwidth]{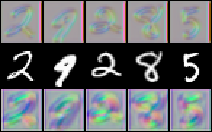}}\hspace{1mm}
\subfigure[SVHN]{\label{fig:svhn_adaptation_target_mnistm}\includegraphics[width=0.21\textwidth]{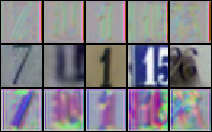}}\hspace{1mm}
\subfigure[SYNTH]{\label{fig:synth_adaptation_target_mnistm}\includegraphics[width=0.21\textwidth]{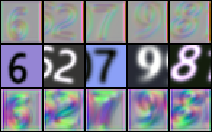}}\hspace{1mm}
\subfigure[MNIST-M]{\label{fig:mnistm_adaptation_target_mnistm}\includegraphics[width=0.21\textwidth]{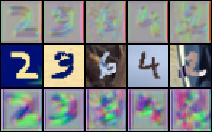}}\hspace{1mm}\vspace{-3mm}
\caption{Examples of domain-agnostic digits generated by  Hallucinator H in the three source experiments with MNIST-M as target.
The top row show images produced in the DG setting by H. The central line shows the original images and in the bottom row we display images produced by H in the DA setting. \textbf{Reminder}: although we can always visualize the \emph{domain agnostic images} to better understand the inner functioning of the network, they are not trained to be be pleasant to the human eye.}\label{fig:examples_best_mnistm}
\end{figure*}
\begin{figure}[t]
\centering
\vspace{-2mm}
\begin{tabular}{@{}c@{~}c@{~}c@{}} 
\includegraphics[width=0.3\textwidth]{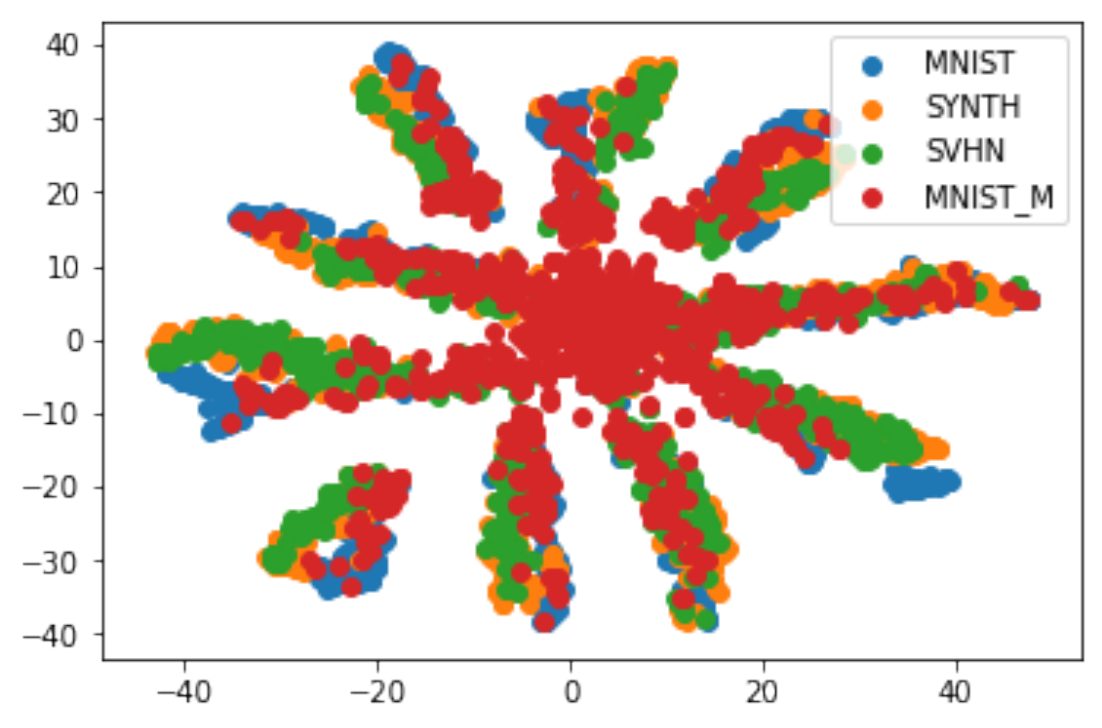} & 
\includegraphics[width=0.3\textwidth]{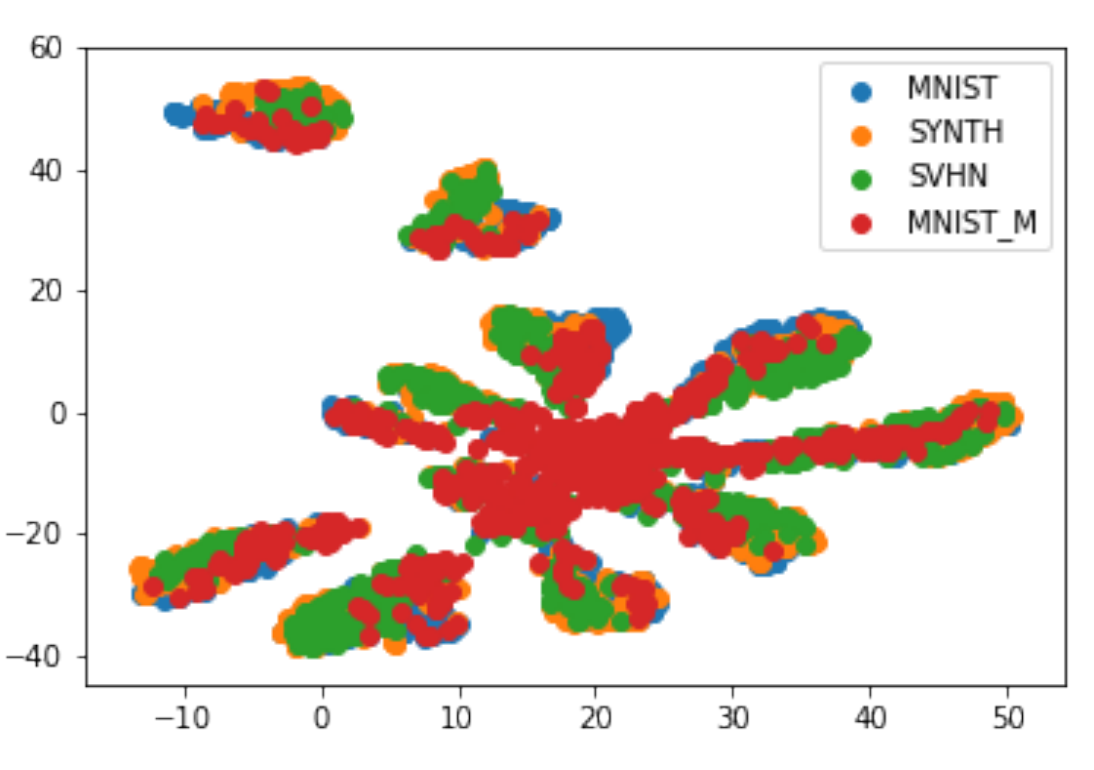} & 
\includegraphics[width=0.3\textwidth]{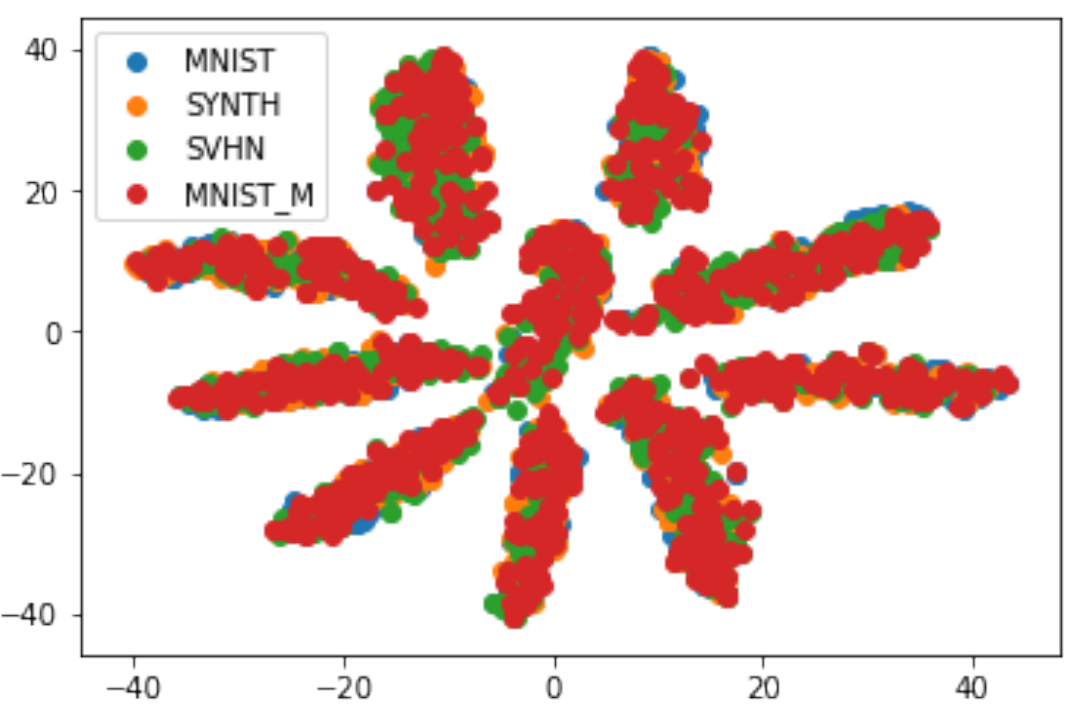} \\
\footnotesize{(a) combine sources} & \footnotesize{(b) DG \our} & \footnotesize{(c) DA \our} \\
\end{tabular}
\caption{TSNE plots of features from the three source experiment with MNIST-M as target.}\label{fig:tsne_mnistm}
\end{figure}

We tested \our\footnote{We implemented \our in Pytorch,  code available at \url{https://github.com/fmcarlucci/ADAGE/}.
} on the DG and multi-source DA scenarios. 
Our framework can easily switch between the two cases with a few key differences. For DG the image $I$ 
and the feature $D$ domain discriminators deal with $S$ domains, while for DA they need to distinguish among
$S+1$ domains including the target. Moreover, in DA, the unlabeled target data trigger the classification
block $C$ to activate the entropy loss and to use the source domain weights provided by the image 
domain discriminator $I$. Specifically these weights make sure that our classifier is biased towards
the sources more similar to the target.

\subsection{Domain Generalization}
\textbf{Datasets} 
We focus on five digits datasets and one object classification dataset. 
\emph{MNIST}~\cite{lecun1998gradient} contains $70$k centered, $28\times28$ pixel, grayscale 
images of single digit numbers on a black background. 
\emph{MNIST-M}~\cite{Ganin:DANN:JMLR16} is a variant where the background is substituted by a 
randomly extracted patch obtained from color photos of BSDS500~\cite{arbelaez2011contour}.
\emph{USPS}~\cite{friedman2001elements} is a digit dataset automatically scanned from envelopes 
by the U.S. Postal Service containing a total of 9,298 $16\times16$ pixel grayscale samples; 
the images are centered, normalized and show a broad range of font styles. 
\emph{SVHN}~\cite{netzer2011reading} is the challenging real-world Street View House 
Number dataset. It contains over $600$k $32\times32$ pixel 
color samples, while we focused on the smaller version of almost $100$k 
cropped digits. Besides presenting a great variety of shapes and textures, images from this dataset 
often contain extraneous numbers in addition to the labeled, centered one.
The Synthetic Digits (\emph{SYNTH}) collection~\cite{Ganin:DANN:JMLR16} 
consists of $500$k images generated from Windows$^{TM}$ fonts by varying the
text (that includes different one-, two-, and three-digit numbers), positioning, orientation,
background and stroke colors, as well as the amount of blur.
Finally, the \emph{ETH80 object dataset} consists of 8 object classes with 10 
instances for each class and 41 different views of each instance with respect to
pose angles. All the images are subsampled to $28\times 28$ and greyscaled.

\begin{table}[!b]
\caption{Domain Generalization accuracy results on experiments with five MNIST-rotated sources. For compactness we only indicate the considered target.}
\label{table:rotatedminst}
{
\resizebox{\textwidth}{!}{
\begin{tabular}{@{~}c@{~}@{~}l@{~}@{~}c@{~}@{~}c@{~}@{~}c@{~}@{~}c@{~}@{~}c@{~}@{~}c@{~}@{~}c@{~}} 
\hline
& Target & $M_{0}$ & $M_{15}$ & $M_{30}$ & $M_{45}$ & $M_{60}$ & $M_{75}$ & Avg.\\ \hline
  \multirow{5}{*}{DG} & D-MTAE \cite{DGautoencoders}   & 82.5 & 96.3 & 93.4 & 78.6 & 94.2 & 80.5 & 87.6\\
& CCSA \cite{doretto2017}   & 84.6 & 95.6 & 94.6 & 82.9 & 94.8 & 82.1 & 89.1 \\ 
& MMD-AAE \cite{Li_2018_CVPR}  & 83.7  & 96.9 & 95.7 & 85.2 & 95.9 & 81.2 & 89.8\\ 
& CROSS-GRAD \cite{DG_ICLR18}  & 88.3 & \textbf{98.6} & \textbf{98.0} & 97.7 & \textbf{97.7} & 91.4 & \textbf{95.3}\\ 
\cline{2-9}
& \our & \textbf{88.8} & 97.6 & 97.5 & \textbf{97.8} & 97.6 & \textbf{91.9} & 95.2\\
\hline
\end{tabular}
}
}
\end{table}

\begin{table}[t]
\centering \footnotesize 
\vspace{-2mm}
\begin{tabular}{@{~}c@{~}@{~}l@{~}@{~}c@{~}@{~}c@{~}@{~}c@{~}@{~}c@{~}@{~}c@{~}@{~}c@{~}} 
\hline
& Target & \scriptsize{$ETH_{00}$} & \scriptsize{$ETH_{22}$} & \scriptsize{$ETH_{45}$} & \scriptsize{$ETH_{68}$} & \scriptsize{$ETH_{90}$}  & Avg.\\ \hline
  \multirow{4}{*}{DG}  
  & combine sources & 70.0 & 93.8 & 96.2 & 98.8 & 81.2 & 88.0\\
  & D-MTAE \cite{DGautoencoders} & - & - & - & - & - & 87.9\\
& MLDG \cite{MLDG_AAA18} & \textbf{70.0} & 85.0 & 95.0 & 97.5 & 73.7 & 84.2\\
\cline{2-8}
& \our & 67.5 & \textbf{95.0} & \textbf{100.0} & \textbf{100.0} & \textbf{88.8} & \textbf{90.2} \\
\hline
\end{tabular}\vspace{1mm}
\hspace{-2mm}\begin{tabular}{@{}c@{~}c@{~}c@{~}c@{}}\scriptsize 
\includegraphics[width=0.25\linewidth]{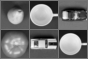} &
\includegraphics[width=0.25\linewidth]{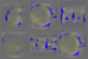} &
\includegraphics[width=0.25\linewidth]{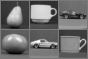} &
\includegraphics[width=0.25\linewidth]{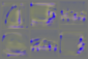}\\
$ETH_{00}$ & hall. $ETH_{00}$& $ETH_{90}$& hall. $ETH_{90}$\\ \hline
\end{tabular} 
\caption{Top: DG accuracy results on experiments with ETH-80 rotated sources. 
Bottom: real and hallucinated image examples.}\label{table:eth80}
\end{table}

\textbf{Scenarios} We consider three experimental scenarios on digits images already presented in previous work.
A first case from \cite{MDAN_ICLRW18} involves \textbf{three sources} chosen in \{MNIST, MNIST-M, SYNTH, SVHN\}.
Each dataset, with the exception of SYNTH, is used in turn as target. All the images are resized to
$28\times28$ pixels and subsets of $20$k and $9$k samples are chosen respectively from each source
and from the target.
A second case from \cite{cocktail_CVPR18} involves \textbf{four sources} by adding USPS to the previous 
dataset group, and focuses on two possible targets, SVHN and MNIST-M. Even in this case the images
are resized to $28\times28$ pixels, and $25$/$9$k samples are drawn from each dataset to define the
source/target sets. 
A third case from \cite{DGautoencoders} involves \textbf{five sources} and exploits rotated variants of MNIST. 
Specifically we started by randomly choosing 100 images for each of the 10 classes and indicating this basic view 
with $M_{0}$. The versions $\{M_{15},M_{30},M_{45},M_{60},M_{75}\}$  are obtained by rotating the images of $15$ 
degrees in counterclock-wise direction.
Note that the authors of \cite{cocktail_CVPR18} kindly shared the exact splits
used for their paper, while for all the other experiments we considered multiple random
selections of the samples from the datasets.
For the \textbf{object classification experiment}, we followed \cite{DGautoencoders} focusing on the ETH80-p setting
that covers 5 domains built from equally spaced pitch-rotated views of the 8 objects. 
Each domain is considered in turn as the target, while the remaining ones are the sources.

\textbf{Implementation Details} For our experiments all the datasets were normalized and zero-centered. 
The mean and standard deviation of the target for data normalization are calculated 
batch-by-batch during the testing process. 
A standard random crop of $90-100\%$ of the total image size was applied as data augmentation.
The training procedure runs for $600$ epochs with Adam optimizer \cite{kingma2014adam} . The initial learning
rate is set to $1e^{-3}$ and step down after $80\%$ of the training.
All the experiments are repeated tree times and we report the average on the obtained classification 
accuracy.

\textbf{Results in Table \ref{table:34sources} (top part)} 
As a main baseline for the three and four sources settings we use the na\"ive \emph{combine sources}
strategy that consists in learning a classifier on all the source data combined together. 
For a fair comparison we produced these results by keeping on only 
the feature extractor $F$ and the classifier $C$, while turning off all the adaptive blocks
in the domain generalizer.
We benchmark against the meta-learning method MLDG \cite{MLDG_AAA18} using
the code provided by the authors and running the experiments on our settings. 
The obtained results indicate that \our 
outperforms all the reference sota baselines in DG both using three and four sources with an advantage up
to 3 percentage points. Interestingly, using four sources slightly worsens the performances when SVHN is the target: 
our interpretation is that adding the USPS dataset 
increases the domain shift between the training and test domains, making the adaptation somehow more difficult.

\textbf{Results in Table \ref{table:rotatedminst}} For the five sources experiments on rotated digit images 
we benchmark against two autoencoder-based DG methods D-MTAE
and MMD-AAE respectively presented in \cite{DGautoencoders} and \cite{Li_2018_CVPR}, as well
as against the metric-learning CCSA method \cite{doretto2017} and the very recent CROSS-GRAD \cite{DG_ICLR18}.
The results indicate that \our outperforms three of the four competitors and has results similar to CROSS-GRAD
which proposes an adaptive solution based on data augmentation that could potentially be combined with \our.

\textbf{Results in Table \ref{table:eth80}} 
For the object classification experiments on ETH80-p, \our obtains an average accuracy of 90.2\% , outperforming D-MTAE \cite{DGautoencoders} and MLDG \cite{MLDG_AAA18}.

\subsection{Domain Adaptation}
We extend our analysis to the multi-source DA setting considering the same three and four scenarios
on digits images described in the previous section. In terms of implementation details, the only difference with 
respect to what already discussed above is that we now have all the unlabeled target samples at training time, so their
mean and standard deviation can be calculated at once. Moreover, for the training process we used 
the RmsProp optimizer \cite{rmsprop}, running for $200$ epochs with initial learning rate of $5e^{-4}$.

\textbf{Results in Table \ref{table:34sources} (bottom part)} We benchmark \our against reference results from 
previous DA works. In particular for the three sources
experiments the comparison is with the Multisource Domain Adversarial Network MDAN~\cite{MDAN_ICLRW18}.
Since this method builds over the DANN algorithm \cite{Ganin:DANN:JMLR16} the result
obtained with DANN applied on the combination of all the sources (combine DANN) is also reported. 
For the four sources experiments the main comparison is instead with the Deep Cocktail Network 
(DCN)~\cite{cocktail_CVPR18}, a recent method able to work even with partial class overlap among the sources.
The results indicate that \our outperforms the competing methods also in this setting with an average advantage up to 11 percentage points.
As a further test we verified the obtained weights assigned by the $I$ network component in the three source
setting: when using MNIST-M as target they converge to $\{0.5, 0.3, 0.2\}$ respectively for MNIST, SVHN, SYNTH,
which sounds reasonable given the visual similarity among the domains.

While \our is specifically tailored for the multi-source settings, we checked its behaviour also
in the case of single source DA with access to unlabeled target data. As a proof of concept experiment, we tested \our using SVHN as source and MNIST as target. 
With the same protocol used in our DA experiments,  we achieve $95.7\%$ accuracy, which is 
on par with the very recent \cite{haeusser17} and better than several others competitive methods \cite{cycada, russo17sbadagan,liu2017unsupervised,saito2017asymmetric}.

\subsection{Ablation Study and Qualitative Results}
Our ablation study analyzes 
the effect of progressively enabling the key components of the domain generalizer alone, and in 
combination with the hallucinator.

\textbf{Results in Table \ref{table:ablation2}} 
We start by evaluating the performance obtained when we do not generate the domain agnostic
samples. In this case the hallucinator $H$ is removed from the network and the original images of all sources
 are fed directly to the domain generalizer. In this case, since we cannot modify the original
images, the only active adaptive component is $D$ that operates on the features. Moreover the classifier can also 
take advantage of the entropy loss (that we indicate with $E$) in the DA setting. 
The results indicate that feature alignment is very helpful for DA but can induce 
confusion in DG with results lower than those of the combine sources baseline. 
Another important result is obtained when only $H$ is enabled and the features
are extracted directly from the generated images with the components $I$ and $D$ off. 
In this case the network is not performing any effort
to align the domains and the final accuracy is just slightly better than the combine sources baseline.
This shows that the advantage of \our is clearly not 
just due to the use of 
a deeper architecture.
Keeping the hallucinator $H$ active together with the $D$ component produces a good advantage in accuracy
but only in the DA setting ($H+D=69.9$). Here adding $E$ provides a further advantage ($H+D+E=82.4$).
Overall the entropy loss appears quite effective in the considered scenario: our intuition is that 
the presence of multiple sources helps reducing the risk that the entropy loss might mislead the classifier. 
The contribution of the image domain discriminator $I$ is negligible by itself and this 
behavior can be explained considering that we backpropagate only a
small part of the $I$ gradient ($\gamma = 0.1\lambda$, see section \ref{sec:architecture}).
However its beneficial effect becomes evident in collaboration with the other network modules:
passing from $H+D+E$ to $H+D+E+I$ implies an improvement in accuracy of at least 4 percentage points in the difficult
DG setting, which shows that the adversarial guidance provided by $I$ on $H$ allows for an image adaptation process 
complementary to the feature adaptation one. Note that, since the image domain discriminator backpropagates only on the hallucinator, it is not possible to test any combination containing $I$ but not $H$.

\begin{table}[!t]
\centering
\resizebox{\textwidth}{!}{
\begin{tabular}{@{~}c@{~}|c@{~~~}c@{~~~}c@{~~~}c@{~~~}c@{~~~}c@{~~~}c@{~~~}c@{~~~}c@{~~~}c@{~~~}c@{~~~}|@{~~}c@{~~}}
\hline
 combine &   &
 \multirow{2}{*}{D} & \multirow{2}{*}{D+E} & \multirow{2}{*}{H} & \multirow{2}{*}{H+E} & \multirow{2}{*}{H+D}  & \multirow{2}{*}{H+I} & \multirow{2}{*}{H+D+I} & \multirow{2}{*}{H+E+I} & \multirow{2}{*}{H+D+E} & \multirow{2}{*}{H+D+E+I} & \multirow{2}{*}{H$_{res}$+D+E+I}\\
 sources &   &  &   &  &  &  &  &  &  &  &   & \\
 \hline
\multirow{2}{*}{62.6} & DG  & 53.0& 53.0 & \multirow{2}{*}{63.2} & 63.2& 62.2& 61.4& 66.3& 61.4& 62.2&66.3 & 65.8\\
                      & DA  & 65.9& 75.1 &                       & 63.9& 69.9& 60.8& 68.8& 63.9& 82.4&88.5 & 87.6\\
\hline
\end{tabular}
}
\caption{Ablation analysis on the experiment with three sources and target MNIST-M.
We turn on and off the different parts of the model: \textbf{H}= Hallucinator, 
\textbf{E}= Entropy, \textbf{D}= Feature Domain Discriminator, \textbf{I}= Image Domain Discriminator. 
Note that H+D+E+I corresponds to our whole method \our.}
\label{table:ablation2}
\vspace{-2mm}
\end{table}

Finally we propose a benchmark against an existing residual structure previously used to transform pixels
in depth image colorization \cite{carlucci2018text}.
When plugging in this residual version of the hallucinator ($H_{res}$) we observe that the overall classification 
performance is slightly lower than what obtained with our original aggregative $H$. 
Besides this small variation, the most important difference is that our hallucinator has only
$1/3$ of the parameters of \cite{carlucci2018text}, thus it is faster in training and allows to 
avoid overfitting while mapping the source domain images into a compact agnostic space.

\textbf{Qualitative Analysis} 
Figure \ref{fig:examples_best_mnistm} shows the agnostic images generated by the hallucinator, in the 
three source experiment with target MNIST-M, while the bottom part of Table \ref{table:eth80} shows examples
of ETH-80 original and hallucinated images.
The main effect of $H$ is that of removing the backgrounds and enhancing the edges:
this is quite clear in the DG setting for both digits and objects, while in the DA case the produced 
digits images appear slightly more confused.
Figure \ref{fig:tsne_mnistm} shows the TSNE embedding of features extracted immediately before the final classifier. 
In the DA setting we completely align the feature spaces of the domains, resulting in a clear per class clustering. 
In the DG setting the results are less clean, but the clusters are still tighter than those obtained by the combine source baseline.

%% file: supplementary.tex

\paragraph{The Hallucinator} In terms of the proposed deep learning architecture, the main technical novelty of our work is in the incremental structure of the Hallucinator module.
A previous work has considered the idea of combining a sequence of consecutive layers by concatenation 
at the end of the network  with a structure known as \emph{hypercolumn} \cite{BharathCVPR2015}, but this approach is different from our Hallucinator: at any point in its structure we have access to all the information of the previous layers which are dynamically stacked. To further confirm the superiority of our $H$ we extended the three sources digits experiments showing what happens when our $H$ is substituted with a simple convolutional, residual or hypecolumn structure. The results are reported in Table \ref{tab:differentH}.

\vspace{-3mm}\paragraph{Image Domain Discriminator and Domain Weights } 
Note that  $I$ and $D$ perform the same task of domain recognition, but having auxiliary losses at different levels in the network is a good strategy to better guide the learning process (\eg see \cite{szegedy2015going}). 
We underline that $I$ is closer to $H$ than $D$, thus $H$ receives a cleaner signal from $I$. At the same time, the source-target domain similarity can be estimated better by $I$ than $D$. We extended the ablation study, turning off the gradient back-propagation from $I$ but keeping the domain similarity weights. We also moved the weights evaluation from $I$ to $D$. Results (Table \ref{tab:ablation}) show that adding the weights over $H+D+E$ provides a small improvement, but 
not enough to outperform \our.

\begin{figure}[ht]
\begin{tabular}{@{}c@{}c@{}c@{}}
\includegraphics[width=0.34\linewidth]{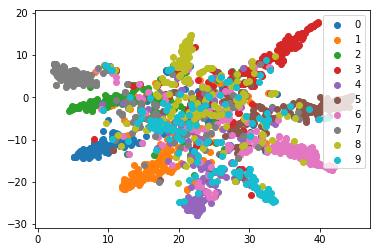}     &  
\includegraphics[width=0.34\linewidth]{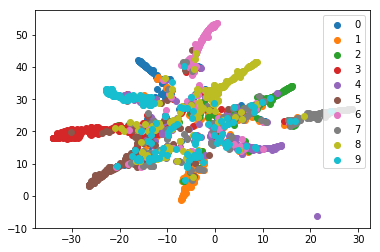} &
\includegraphics[width=0.34\linewidth]{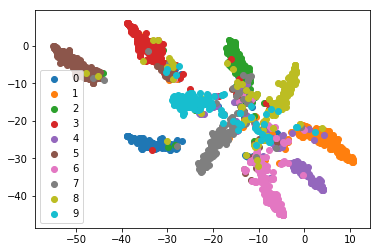}\\
\footnotesize{(a) combine sources} & \footnotesize{(b) DG ADAGE}   &  \footnotesize{(c) DA ADAGE} 
\end{tabular}
\caption{TSNE visualization: same setting of Fig. 5 in the main submission. Here we show only the MNIST-M target with each sample labeled according to its class.}
\label{fig:tsne_mnistm_appendix}
\vspace{2mm}
\end{figure}

\vspace{-3mm}\paragraph{Target Visualization} In Figure \ref{fig:tsne_mnistm} we showed two-dimensional TSNE visualization for the network features by using colors to differentiate among the domains. Here we repeat the exercise by using colors to indicate the sample class labels. The plots in Figure \ref{fig:tsne_mnistm_appendix} indicate that, although the entropy loss promotes a small degree of class confusion, \our clearly leads to discriminative features.

\setlength{\tabcolsep}{10pt} 
\begin{table}[t]\centering \footnotesize
\begingroup
\renewcommand{\arraystretch}{1} 
\begin{tabular}{@{}l@{~}l@{}c@{}c@{}c@{}c}  
\hline
   \multicolumn{2}{c}{ \multirow{3}{*}{Sources}}& SVHN & SVHN & MNIST-M&\multirow{4}{*}{Avg.}\\ 
  \multicolumn{2}{c}{ }& MNIST-M & MNIST & SYNTH\\ 
   \multicolumn{2}{c}{ }& SYNTH & SYNTH & MNIST\\ \cline{2-5} 
  \multicolumn{2}{c}{Target} & MNIST & MNIST-M & SVHN\\ \hline
\multirow{4}{*}{DG} & $H$ Incremental (\footnotesize{\our})          &  99.1  & 66.3 & \textbf{76.4}  & \textbf{80.3}\\
& $H$ Convolutional    &  97.9    & 64.0   &  69.5  & 77.1 \\
 & $H$ Residual \cite{carlucci2018text}           &  \textbf{99.2}  & 65.8 & 74.6 & 79.9 \\
  & $H$ Hypercolumn \cite{BharathCVPR2015}   &   97.7   &  \textbf{68.0}  & 70.5   & 78.7  \\
 \hline \hline
\multirow{4}{*}{DA}   & $H$ incremental (\footnotesize{\our})     &  \textbf{99.3}   &  \textbf{88.5} &  \textbf{86.0} & \textbf{91.3}\\
& $H$ Convolutional     &  99.1    & 88.2   & 84.7   & 90.6 \\
  & $H$ Residual \cite{carlucci2018text}   &  99.2   &  87.6 &  84.1 & 90.3\\
  & $H$ Hypercolumn \cite{BharathCVPR2015}    &   99.0   & 80.0   &  83.5  & 84.5 \\
\hline
\end{tabular}
\endgroup
\caption{Comparison of the incremental $H$ architecture used in \our with respect to other possible variants.} \label{tab:differentH}
\end{table}
\begin{table}
\begin{tabular}{@{}l@{}c@{}}
\hline
\multicolumn{2}{@{}c@{}}{ \{SVHN MNIST SYNTH\} $\rightarrow$ MNIST-M}\\
\hline
$H+D+E$ & 82.4\\
$H+D+E+$weights($I$) & 84.5\\ 
$H+D+E+$weights($D$) & 83.3 \\
$H+D+E+I$ (\footnotesize{\our}) &  \textbf{88.5}\\
\hline
\end{tabular}
\caption{Extended ablation study on the DA case. In DG the weights are always off.}
\label{tab:ablation}

\end{table}%

\paragraph{Discussion about Related Works}
While \our is intuitively similar to \cite{carlucci2018text} and \cite{GatysEB16},
 commonalities end at a high level description. 
The goal of \cite{carlucci2018text} is to colorize 
raw depth images  
to optimally feed them to an RGB pretrained model. Most of the network is frozen and the objective is to minimize a cross-entropy classification loss. \our maps images to a domain agnostic space by minimizing two domain confusion losses besides the cross-entropy object classification loss, with the \emph{whole} network trained end-to-end. Moreover the incremental structure of the $H$ block is substantially different from the residual architecture used in [7]. Quantitative evidence is provided 
in Table \ref{tab:differentH}. 

The work \cite{GatysEB16}
assumes that the image style is captured by the correlations between the different filter responses of any layer of the network, supposing that the style is specific for each image. \our lets the network free to learn what should be considered as style over multiple images through domain confusion conditions. We make no assumptions neither on which specific part of the network captures the style, nor on how to discard it: both are automatically optimized while training 
across multiple domains. Indeed, by avoiding to define the style separately for any image we make it possible to deal jointly with multiple sources.